\documentclass[letter, 10pt, conference]{ieeeconf}

\IEEEoverridecommandlockouts                              
\usepackage{amsmath}
\usepackage{cite}
\usepackage{svg}
\usepackage{graphicx}
\usepackage{algorithm,algpseudocode}
\usepackage{caption, subcaption}
\usepackage{rotating}
\usepackage{multirow}
\usepackage{array}
\usepackage{hyperref} 
\usepackage{amsmath}
\usepackage{amssymb}
\usepackage{graphicx}
\usepackage{algorithm}
\usepackage{algpseudocode}
\usepackage{epstopdf}
\usepackage{mathtools}
\usepackage{makecell}
\usepackage{fancyhdr}
\usepackage{soul}
\usepackage{slashbox}
\usepackage{hhline}
\usepackage{pifont}
\usepackage{fancyhdr}
\usepackage{float} 

\title{\LARGE \bf
Context-Aware Risk Estimation in Home Environments: A Probabilistic Framework for Service Robots}

\author{
Sena Ishii, Akash Chikhalikar, Ankit A. Ravankar, Jose Victorio Salazar Luces, and Yasuhisa Hirata%
\thanks{All authors are with the Department of Robotics, Graduate School of Engineering, Tohoku University, Sendai 980-8579, Japan. Email: \{sena.ishii, a.k.chikhalikar, ankit, j.salazar, hirata\}@srd.mech.tohoku.ac.jp}}

\renewcommand{\baselinestretch}{0.959}
\fancypagestyle{firstpage}{%
  \fancyhf{} 
  \fancyfoot[C]{\footnotesize
    This work has been submitted to the IEEE for possible publication.
    Copyright may be transferred without notice, after which this version
    may no longer be accessible. This work has been accepted for the 34th IEEE International Conference on Robot and Human Interactive Communication, IEEE RO-MAN 2025.}%
}

\begin{document}

\maketitle
\thispagestyle{firstpage}
\pagestyle{empty}

\begin{abstract}
We present a novel framework for estimating accident-prone regions in everyday indoor scenes, aimed at improving real-time risk awareness in service robots operating in human-centric environments. As robots become integrated into daily life, particularly in homes, the ability to anticipate and respond to environmental hazards is crucial for ensuring user safety, trust, and effective human-robot interaction. Our approach models object-level risk and context through a semantic graph-based propagation algorithm. Each object is represented as a node with an associated risk score, and risk propagates asymmetrically from high-risk to low-risk objects based on spatial proximity and accident relationship. This enables the robot to infer potential hazards even when they are not explicitly visible or labeled. Designed for interpretability and lightweight onboard deployment, our method is validated on a dataset with human-annotated risk regions, achieving a binary risk detection accuracy of 75\%. The system demonstrates strong alignment with human perception, particularly in scenes involving sharp or unstable objects. These results underline the potential of context-aware risk reasoning to enhance robotic scene understanding and proactive safety behaviors in shared human-robot spaces. This framework could serve as a foundation for future systems that make context-driven safety decisions, provide real-time alerts, or autonomously assist users in avoiding or mitigating hazards within home environments.

\end{abstract}


\section{Introduction}

\begin{figure}[!ht]
    \centering
    \includegraphics[width=0.45\textwidth]{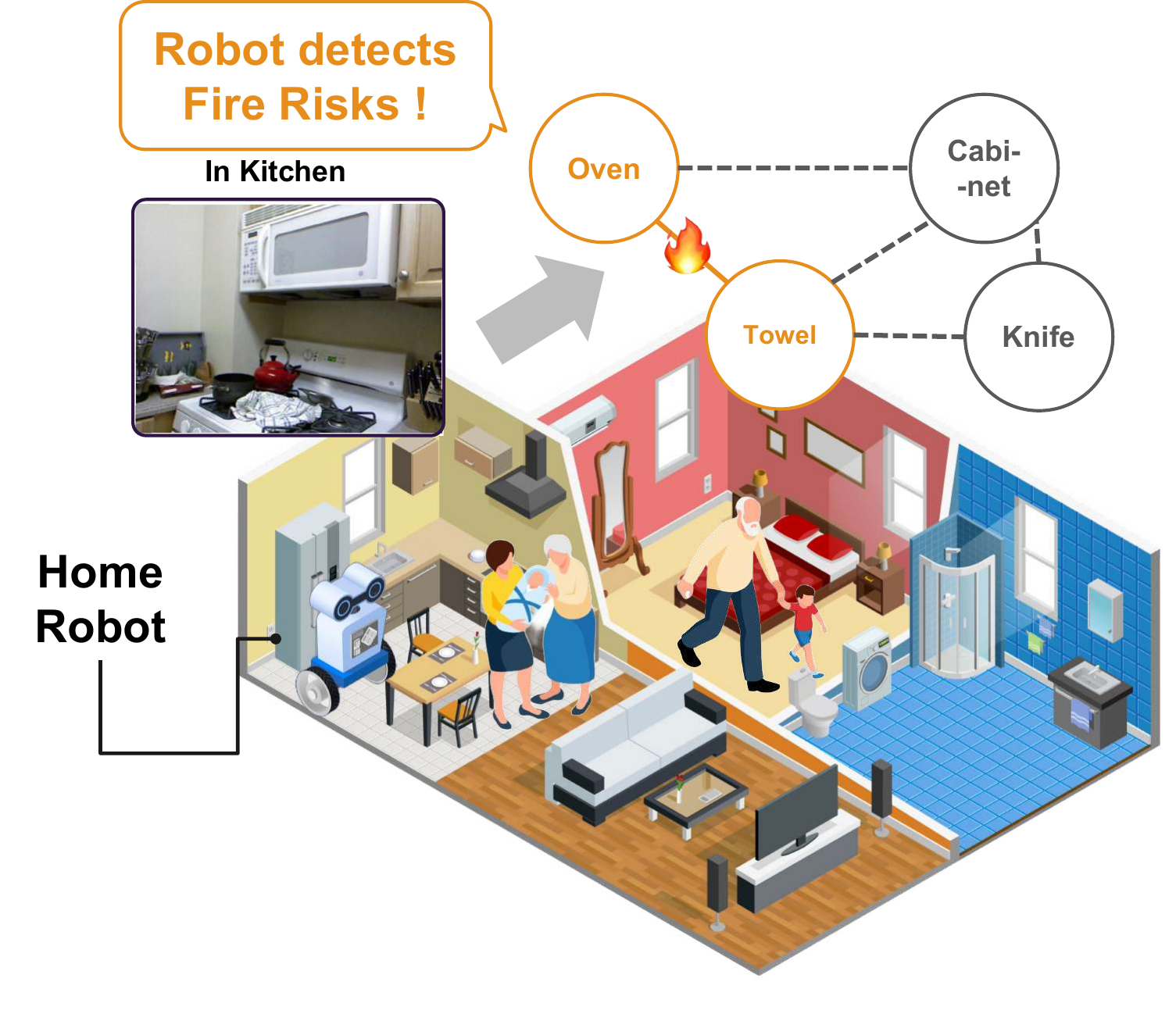}
    \caption{Overview of the proposed risk propagation framework in a household environment. 
The home robot detects potential fire risks by analyzing the spatial proximity and accident-related correlations among objects. 
In this scenario, the towel near the oven was initially not recognized as hazardous, 
but after risk propagation, it is identified as a high-risk object due to its proximity to the oven.}
    \label{fig:algorithm}
\end{figure}


As service robots become increasingly integrated into daily life—supporting tasks such as cleaning, acting as communication companions, searching for objects  or navigating shared spaces~\cite{black2024pi0, 9976330, chikhalikar2025enhancingobjectsearchindoor}—their roles are expected to expand beyond single-function behaviors. With recent advances in embodied intelligence and large language models, these robots are beginning to understand complex instructions and act autonomously in diverse home environments.

While recent advancements in domestic robotics have led to increasingly sophisticated capabilities the development of systems that can anticipate and mitigate household accidents remains equally critical.  Building user trust is paramount for the widespread adoption of service robots; a robot capable of proactively identifying and mitigating risks not only ensures physical safety but also fosters confidence and acceptance among users, particularly in vulnerable populations.

This need is especially pronounced in aging societies, where in-home safety is a pressing concern~\cite{MHLW2013, MHLW2023}.
Service robots capable of predicting and responding to potential risks can play a vital role: not only by taking preemptive actions to prevent accidents, but also by providing timely responses when incidents occur. This function becomes particularly meaningful when integrated into mobile home robots, whose mobility and environmental awareness make them promising platforms for augmenting with secondary functions such as real-time risk-aware behavior. By equipping these robots with contextual risk reasoning capabilities, we can expand their utility beyond their original purpose and move toward safer, more adaptive home environments.

In this context, unlike conventional robotic systems focused primarily on geometry or semantic recognition\cite{AKASHLATEST}, service robots must go a step further: they must anticipate \emph{contextual risk} in their surroundings~\cite{haddadin2017physical, alem2020real}.

There are existing robotic technologies aimed at predicting household accidents. For instance, Xiuying Gen et al. (AnomalyGen)~\cite{song2024hazards} proposed a framework that leverages large language models and 3D simulation environments to identify and avoid potential hazards in home settings. However, this approach solely determines risks based on observed objects, sometimes misidentifying non-hazardous situations as risky. For example, a knife placed on a counter is not inherently dangerous unless positioned near an edge or within reach of a child. Similarly, an entangled cable or an unstable stool may become accident-prone depending on nearby objects or movement in the scene. These risks emerge not just from the objects themselves, but from their \emph{semantic roles, spatial relationships, and affordances}—something humans intuitively recognize, but most robots do not.

Prior works have addressed risk-aware navigation in rough terrain~\cite{trg2024pathplanning}, fall-risk screening in eldercare~\cite{georgiev2021fall}, and unsafe event detection in surgical robotics~\cite{alem2020real}. Other approaches rely on scene graphs~\cite{dellaert2017factor}, collaborative graph neural networks~\cite{gnn2022collab}, or foundation models for task planning~\cite{AkashAccess,zhu2024earbench,chikhalikar2023integrating} to model object relationships and improve safety perception. However, these methods either: (i) focus on \emph{traversability} or physical terrain, (ii) require \emph{static rules} or explicit task instructions, or (iii) lack real-time inference of \emph{latent risk} in dynamic, unstructured home scenes.

Our work addresses this gap by focusing on \emph{visual risk perception in everyday indoor environments}, where hazards often depend on subtle object interactions and context. We propose a lightweight, onboard-friendly framework that integrates: (1) object-level risk estimation, (2) semantic-spatial relationships, and (3) asymmetric risk propagation through a scene graph.

Unlike prior methods, our system does not depend on large symbolic models or handcrafted rules, and can infer indirect risks—such as a bowl near the edge of a shelf stacked above a slippery surface—by interpreting the scene holistically. 

To reduce false positives and avoid excessive detection, a risk inference system that dynamically adjusts risk values based on environmental changes and personalizes risk assessments for different households is essential. By incorporating multiple parameters into the risk estimation process and formalizing the evaluation, a flexible and adaptable risk reasoning system can be constructed. In this study, we propose a foundational algorithm for accident risk prediction in home environments to enhance robot-assisted safety monitoring. Our approach enables robots to not only recognize immediate hazards but also adapt to diverse and evolving domestic contexts.

We validate our method using a dataset with human-annotated risk regions, achieving 75\% binary risk detection accuracy. Importantly, we demonstrate that our model's predictions closely align with human intuition, especially in scenes involving unstable placements, sharp tools, or cluttered environments.

\textbf{The key contributions of this paper are:}
\begin{itemize}
    \item A novel semantic graph-based framework for estimating accident-prone regions using contextual risk propagation.
    \item An asymmetric risk diffusion algorithm that models how risk flows between semantically related and spatially proximate objects.
    \item Empirical validation against human-labeled data, demonstrating strong alignment with human perception of risk.
    \item A lightweight design suitable for real-time onboard deployment in assistive or mobile robots operating in homes.
\end{itemize}

This work represents a step toward socially aware, perceptive robots capable of understanding risk not just as geometry, but as semantics-in-context, enabling safer and more intuitive human-robot coexistence.


\section{Related Work}

Recent advancements in robotics and assistive technologies have driven research toward improving safety and quality of life for elderly individuals and families with young children. With robots increasingly integrated into domestic spaces, safety-aware perception systems are becoming essential components of home robotics\cite{hirata2022cooperation,cocsar2020enrichme,wang2019novel}.

Several recent approaches have explored robot-assisted safety in indoor environments. AnomalyGen~\cite{song2024hazards} combines large language models with 3D simulations to detect and avoid hazardous scenarios in homes. However, such systems often rely on static object-level detection, which may lead to high false positive rates, particularly when context is ignored. This underscores the need for a dynamic and context-aware risk inference system that accounts for environmental interactions and the spatial arrangement of objects—an area that remains underexplored.

The SafetyDetect dataset, introduced by Mullen et al.~\cite{10601323}, proposes a household anomaly detection framework using scene graph representations and large language models (LLMs) like GPT-4. Their system infers risk based on textual relationships between objects derived from image annotations. While this method demonstrates promising semantic reasoning, it does not directly analyze visual input and depends on intermediate textual scene graph conversion. This reliance may result in loss of visual details critical for subtle risk interpretation.

In parallel, other efforts focus on improving visual grounding and multi-modal information fusion. Chen et al.~\cite{chen2023contextawarefusion} propose a deep-learning-based framework for multi-level information fusion in indoor mobile robots to improve obstacle detection and hazard avoidance, even under occlusions. Similarly, Zhu et al.~\cite{zhu2024earbench} introduce EARBench, a benchmark designed to evaluate physical risk awareness in embodied AI agents. Their findings emphasize the shortcomings of current embodied systems in recognizing and reacting to real-world risks, highlighting the demand for more grounded and interpretable solutions.

Beyond mere detection accuracy, the interpretability of risk assessment is crucial for human-robot collaboration, allowing users to understand why a robot perceives a situation as risky, thereby enhancing trust and enabling more effective joint decision-making.


In contrast, our approach operates directly on \textit{visual data} (RGB-D images) to detect objects and infer \textit{accident risks}. By bypassing the intermediate text representation stage, our system achieves a more direct and robust form of risk assessment, capable of capturing nuanced spatial-semantic cues. Table~\ref{tab:comparison} shows key comparison of existing approaches in household accident risk assessment with other state-of-the art methods.

Furthermore, unlike prior approaches that rely primarily on language models or symbolic reasoning, we incorporate \textit{real-world accident statistics} to assign data-driven risk scores to objects and their relationships. By referencing empirical accident data, our system better aligns with how humans perceive risk, allowing for more intuitive and contextually grounded estimation suitable for mobile robots operating in home environments. Table~\ref{tab:comparison} compares our method against prior approaches in household accident risk prediction based on key capabilities: use of visual input, ability to adjust risk contextually, incorporation of semantic reasoning, and use of real-world data.

\begin{table}[t]
    \centering
    \caption{Comparison of existing approaches in household accident risk assessment.}
    \resizebox{\columnwidth}{!}{
    \begin{tabular}{lcccc}
        \hline
        \textbf{Method} & \textbf{Visual Input} & \textbf{Risk Adjustment} & \textbf{Context Awareness} & \textbf{Real-world Data} \\
        \hline
        AnomalyGen~\cite{song2024hazards} & \ding{51} & \ding{55} & \ding{55} & \ding{51} \\
        SafetyDetect~\cite{10601323} & \ding{55} & \ding{51} & \ding{55} & \ding{51} \\
        EARBench~\cite{zhu2024earbench} & \ding{51} & \ding{55} & \ding{51} & \ding{55} \\
        \textbf{Ours} & \ding{51} & \ding{51} & \ding{51} & \ding{51} \\
        \hline
    \end{tabular}
    }
    \label{tab:comparison}
\end{table}
\section{Methodology}
In this section we present the framework of our risk propagation mechanism based on empirical dataset. 
\subsection{Risk Estimation}

We introduce the concept of a \textit{Risk Score} to quantitatively evaluate the likelihood of household accidents. This score serves as an indicator of how prone a specific object is to causing a particular type of accident. It is computed using real-world statistical data, providing a robust and objective baseline for our model.
\begin{figure}[t]
    \centering
    \includegraphics[width=0.45\textwidth]{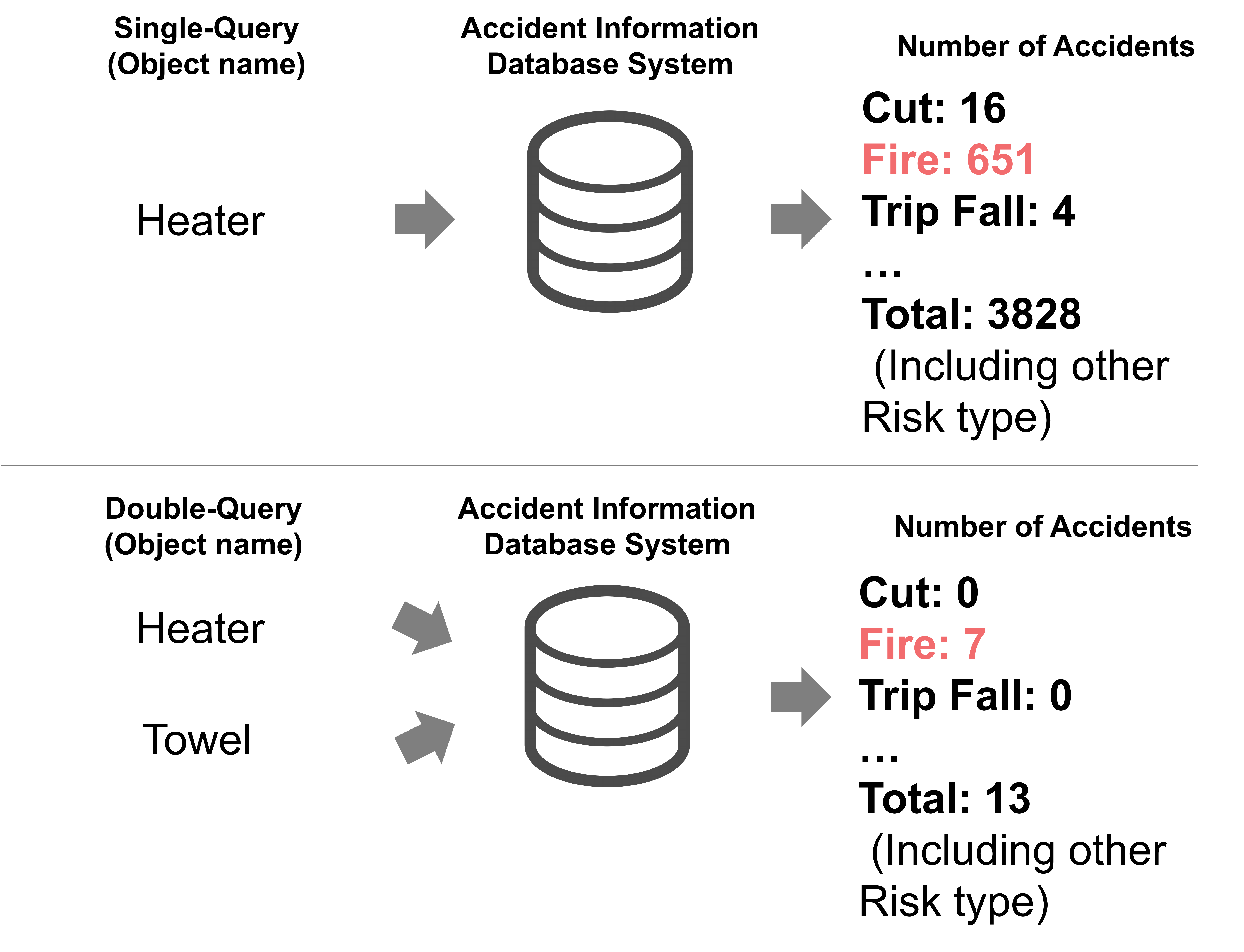}
    \caption{
Visualization of querying the Accident Information Database System for retrieving statistical accident records. 
The upper panel shows a single-object query (e.g., \textit{Heater}) that returns accident counts for each risk type.
The lower panel shows a co-occurrence query for a pair of objects (Towel and Heater), which retrieves accident statistics where both objects were involved in the same incident report. These statistics are used to quantify contextual risk correlations between object pairs, contributing to the propagation component of our framework.
}
    \label{fig:search}
\end{figure}

 \begin{figure*}[tbh]
    \centering
    \includegraphics[width=\textwidth]{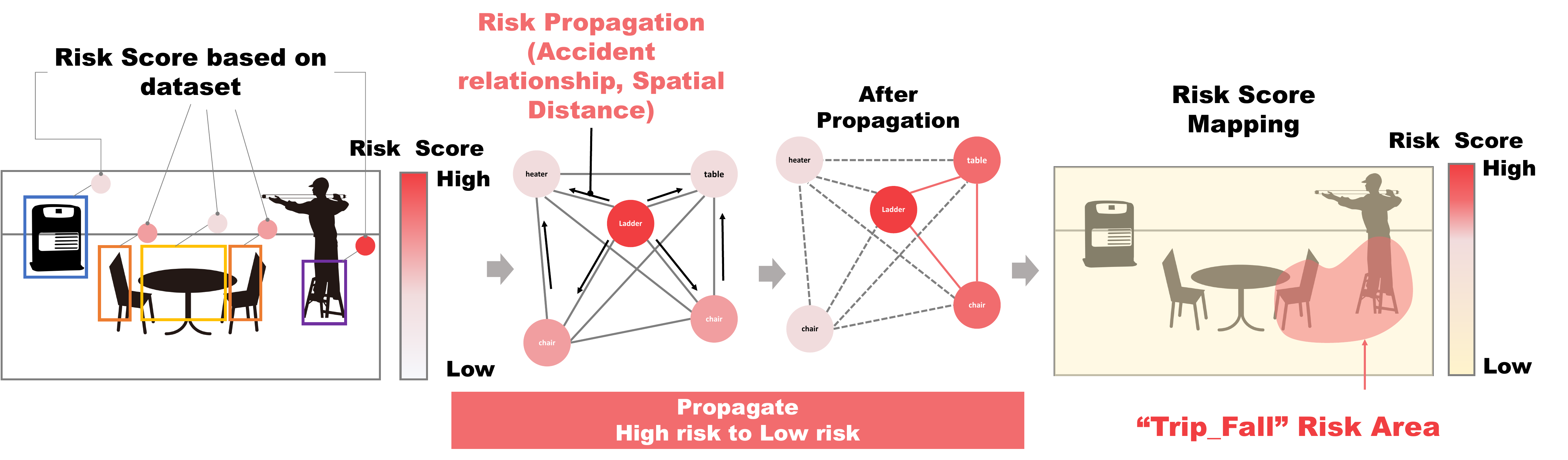}
    \caption{Illustration of risk propagation in a household environment. 
The left section represents the initial risk scores based on dataset priors, where objects are assigned risk values before propagation.
The center graph shows how risk information propagates through spatial distance and dynamically updating risk scores. 
The right section visualizes the final risk map, highlighting high-risk areas such as the ladder, where trip/fall risks are amplified.}
    \label{fig:algorithm}
\end{figure*}
\begin{figure*}[tb]
    \centering
    \includegraphics[width=\textwidth]{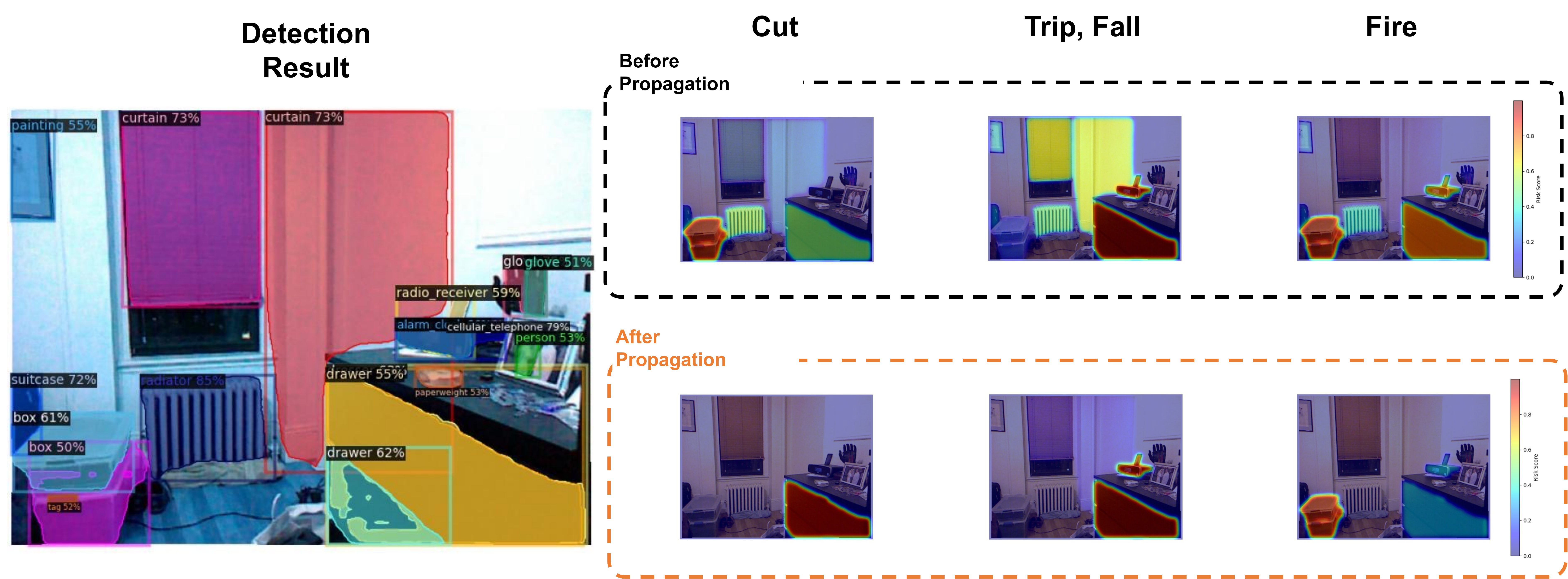}
    \caption{Comparison of risk score maps before and after risk propagation for different accident types: Cut, Trip/Fall, and Fire. 
The top row represents the initial risk distribution, while the bottom row shows the risk scores after propagation.}
    \label{fig:concept}
\end{figure*}

To derive these scores, we use the Accident Information Database System\cite{jikojoho} provided by the Consumer Affairs Agency and the National Consumer Affairs Center of Japan. This web-based database contains approximately 400,000 reports of actual accidents, detailing object involvement, accident causes, and damage outcomes. As shown in Figure~\ref{fig:search}, we query the database using both single-object terms (e.g., \textit{Heater}) and object-pair combinations (e.g., \textit{Towel near Heater}). The single-object queries inform individual risk scores, while pairwise queries capture contextual accident correlations between co-occurring objects.

The risk score $R(o, a)$ for a given object $o$ and accident type $a$ is initially defined using a raw ratio:
\begin{equation}  
R_{\text{raw}}(o, a) = \frac{\text{count}(o, a)}{\text{total}(o)}
\end{equation}

where $\text{count}(o, a)$ denotes the number of accidents of type $a$ involving object $o$, and $\text{total}(o)$ is the total number of accident reports involving $o$. However, this raw ratio may become unstable for objects with few reports, leading to inflated or skewed scores.

To counter this, we apply Laplace-style Bayesian smoothing:

\begin{equation}
R(o, a) = \frac{\text{count}(o, a) + k}{\text{total}(o) + k \cdot N}
\end{equation}

Here, $N$ is the number of accident types, and $k$ is a smoothing parameter (e.g., $k = 1$). This formulation ensures stability for low-frequency objects while preserving accuracy when data is sufficient.

For example, the cut-injury risk for a kitchen knife is calculated as:

\begin{equation}
R(\text{knife, cut}) = \frac{N(\text{knife, cut})}{N(\text{knife})}
\end{equation}

This scoring method supports the systematic evaluation of object risk for specific accident categories. In this study, we focus on three types:
\begin{itemize}
    \item Cut injuries
    \item Fire-related accidents
    \item Falls
\end{itemize}

\subsection{Risk Propagation Algorithm}

In many household scenarios, the danger associated with an object arises not only from the object itself but also from how it interacts with nearby items. For example, while a towel is relatively harmless on its own, its placement near a stove significantly increases the risk of fire. To mimic the human intuition for such context-sensitive hazards, we propose an asymmetric Risk Propagation Algorithm based on an object-centric scene graph.

Each object in the scene is represented as a node with an initial risk score. These scores are iteratively updated by accounting for both semantic relationships and spatial proximity between neighboring objects.




\paragraph{Accident-Related Correlation ($\phi_\text{accrel}$)}
Quantifies how strongly two objects interact to cause accidents. For example, a stuffed toy on its own is generally not associated with a high risk of accidents. However, if it is placed on top of a stove, the risk of fire increases significantly. This concept is introduced to capture such context-dependent interactions between multiple objects, where the likelihood of an accident changes based on their spatial and functional relationships.
\begin{equation}
\phi_{\text{accrel}}(o_1, o_2, a) = \frac{\text{count}(o_1, o_2, a) + k}{\text{total}(o_1, o_2) + k \cdot N}
\end{equation}

\paragraph*{Example.}
Assume the following statistics from the database:
\begin{itemize}
    \item Object pair: \textit{Towel} and \textit{Stove}
    \item Accident type: \textit{Fire}
    \item $\text{count}(Towel, Stove, Fire) = 6$
    \item $\text{total}(Towel, Stove) = 10$
    \item $N = 3$, $k = 1$
\end{itemize}

\begin{equation}
\phi_{\text{accrel}}(\text{Towel}, \text{Stove}, \text{Fire}) = \frac{6 + 1}{10 + 1 \cdot 3} = \frac{7}{13} \approx 0.538
\end{equation}

This value indicates a moderate accident correlation between a towel and a stove in the context of fire, thereby increasing the risk propagation effect to nearby objects.

\paragraph{Spatial Distance ($\phi_\text{distance}$)}
Represents the physical proximity between two objects. It is computed using the Euclidean distances between object centroids in 3D space. Closer objects exert stronger influence on each other, consistent with real-world risk factors (e.g., a curtain placed near a heater).

\paragraph{Risk Difference ($\max(0, r_j - r_i)$)}
Ensures that risk only propagates from high-risk to lower-risk objects, reflecting the asymmetry in real-world risk influence. This prevents inflation of risk from benign objects and promotes selective propagation.

\subsection{Algorithm Description}
We model risk propagation in an object-centric scene graph by simulating how humans intuitively perceive and transfer risk. At each iteration, the risk score of an object is updated based on the influence of its neighbors. The algorithm allows for an asymmetric flow of risk—from objects with higher risk to those with lower risk.

For a node $i$ and its neighbor $j$, the propagation weight is:
\begin{align}
w_{ij} = \phi_\text{accrel}[i,j] \cdot (1 - \phi_\text{distance}[i,j]) \cdot \max(0, r_j - r_i)
\end{align}
Here, $r_i$ and $r_j$ denote the current risk values of object $i$ and $j$, respectively. Risk propagates only when $r_j > r_i$, enforcing a unidirectional flow from higher-risk to lower-risk entities. 
This asymmetric design reflects the intuition that dangerous objects influence their surroundings, but not vice versa.

The factor $\phi_\text{accrel}[i,j]$ captures semantic accident relevance, while $(1 - \phi_\text{distance}[i,j])$ modulates influence by spatial proximity, ensuring that closer objects have stronger impact(e.g., “Clothes placed in close proximity to a heater can increase the risk of fire-related accidents.”). 
This encourages localized propagation, reflecting the fact that physical closeness typically increases accident likelihood.

\begin{algorithm}[!t]

\caption{Asymmetric Risk Propagation}
\begin{algorithmic}[1]
\Require Scene graph $G = (V, E)$ with risk score $r_i$ for each node $i$
\State Normalize all $r_i$ to $[0,1]$
\For{each iteration up to \texttt{max\_iterations}}
    \State $max\_diff \gets 0$
    \For{each node $i \in V$}
        \State $influence \gets 0$, $total\_weight \gets 0$
        \For{each neighbor $j$ of $i$}
            \If{$r_j > r_i$}
                \State $w_{ij} \gets \phi_{\mathrm{accrel}}[i,j] \cdot (1 - \phi_{\mathrm{distance}}[i,j])$
                \State $influence \gets influence + w_{ij} \cdot (r_j - r_i)$
                \State $total\_weight \gets total\_weight + |w_{ij}|$
            \EndIf
        \EndFor
        \State $\Delta r_i \gets \frac{influence}{total\_weight + \epsilon}$
        \State $r_i \gets \text{clip}(r_i + \Delta r_i, 0, 1)$
        \State Update $max\_diff$ if $|\Delta r_i|$ is larger
    \EndFor
    \State Re-normalize all $r_i$ using min-max scaling
    \If{$max\_diff < \texttt{tolerance}$}
        \State \textbf{break}
    \EndIf
\EndFor
\end{algorithmic}
\textit{Note:} The function $\text{clip}(x,\ 0,\ 1)$ limits the value to the range $[0, 1]$ to ensure valid probabilities.
\end{algorithm}

As shown in Fig.~\ref{fig:algorithm}, the process begins with analyzing RGB-D images to detect objects and obtain their three-dimensional positions. Object detection is performed using Detic~\cite{zhou2022detecting}, and the centroid coordinates of the bounding box are used as the object's location. Each object is initially assigned a risk score $R(o, a)$, and a scene graph is constructed using spatial and semantic edges.

To spatially visualize risk scores, heatmaps are generated over RGB images using object detection results and segmentation masks. The process involves:
\begin{itemize}
    \item Assigning risk scores to pixels within the detected object masks.
    \item Applying Gaussian filters (\( \sigma = 3 \)) to smooth abrupt transitions between high-risk and low-risk areas.
    \item In the heatmap visualization, red indicates high-risk areas, while blue denotes low-risk areas.
    \item Normalizing risk scores to the range \( [0, 1] \) for interpretability.
\end{itemize}

In the Fire category, the scenario presented in Figure \ref{fig:concept} highlights a critical refinement by our propagation model. Initially, the box near the radiator correctly receives a high fire risk score, as depicted in the top row. However, the cabinet in the foreground is also erroneously attributed a high risk. Ideally, only the box and radiator should be highlighted for fire risk in this context. Through risk propagation, shown in the bottom row, the cabinet's risk score is effectively reduced, and the high fire risk is appropriately highlighted for the box only. This demonstrates our model's selectivity and effectiveness in correcting initial over-estimations. Conversely, in Trip/Fall scenarios where strong contextual correlations are absent, risk scores remain largely unchanged, further emphasizing the model's selective propagation.

\section{Experimental Results}
\begin{figure}[t]
\centering
\includegraphics[width=0.45\textwidth]{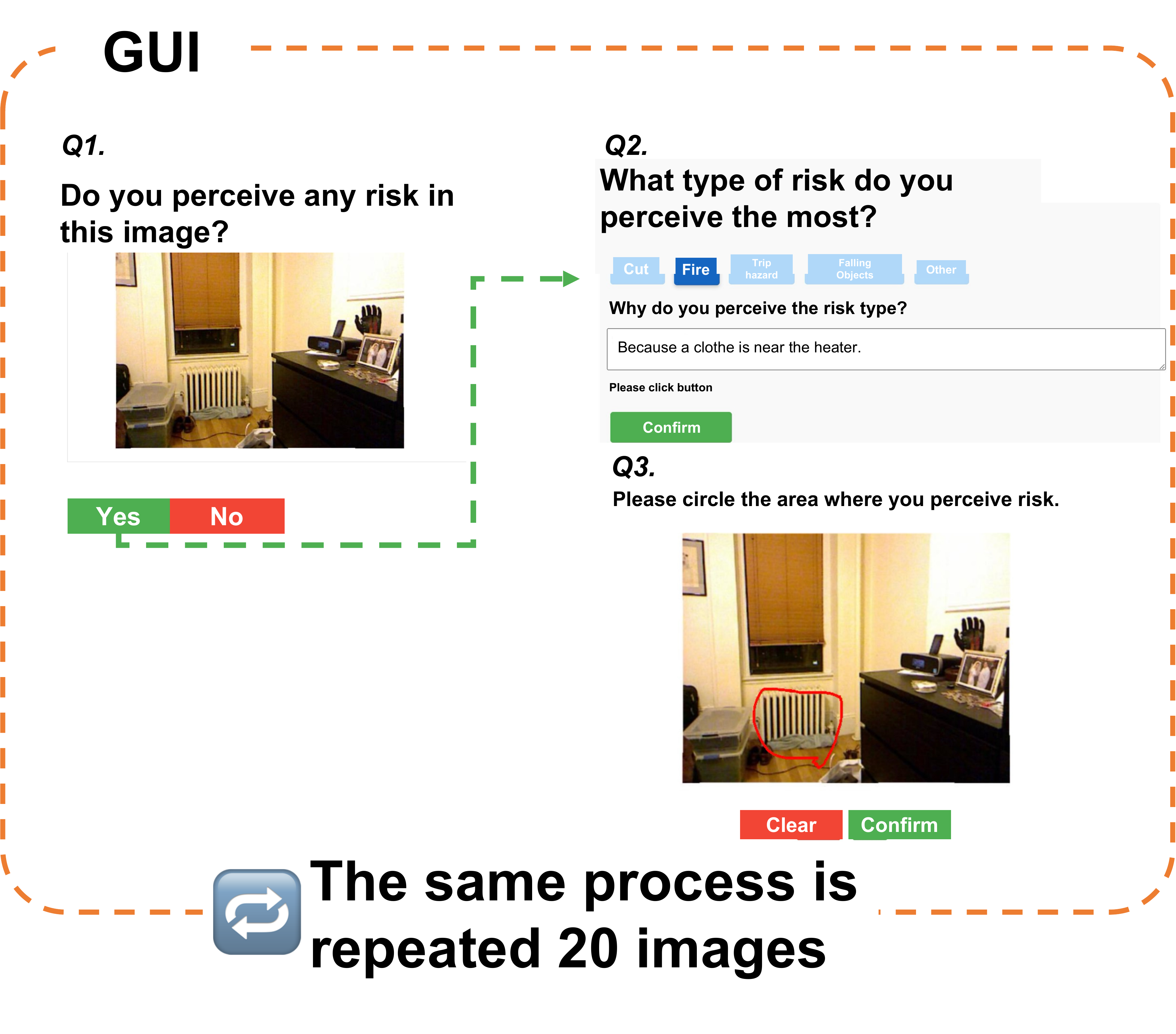}  
\caption{Evaluation GUI Flow}
\label{fig:gui_flow}
\end{figure}

To validate the effectiveness of our proposed risk propagation algorithm, we conducted a two-stage user study focused on understanding how well the system aligns with human perception in detecting and classifying household accident risks.

While the system is designed for deployment on real-time robotic platforms equipped with RGB-D sensors, this initial evaluation was deliberately carried out using the NYU Depth V2 dataset~\cite{Silberman:ECCV12}. This publicly available dataset provides diverse, high-resolution RGB-D imagery of real domestic scenes, making it ideal for benchmarking risk perception without introducing the variability and noise inherent in robot-collected data. Our goal at this stage was to validate the core reasoning capabilities of the system in a controlled and well-understood setting.

The study aimed to address two key research questions:
\begin{itemize}
   
\item Can the system accurately identify whether a household scene contains potential accident risks?
\item Can it correctly classify the identified risk into one of the following categories: cut, fire, or tirp/fall?

\end{itemize}
To collect reference data for evaluation, we developed a web-based GUI (Fig.~\ref{fig:gui_flow}) that presented 20 RGB-D images from the NYU dataset to human participants. For each image, participants were asked to:
\begin{itemize}
	\item Indicate whether they perceived any accident risk.
	\item Choose the type of risk (cut, fire, or fall).
	\item Annotate specific regions in the image that corresponded to the perceived hazard. 
\end{itemize}

This setup enabled us to capture both spatial and semantic information from human annotators. To minimize bias, we recruited 14 participants, including individuals not affiliated with the research team. Participants received only minimal instructions and were encouraged to rely on intuition. Before beginning the formal task, each participant completed a short practice session using four sample images to familiarize themselves with the interface.

Each RGB-D image was subsequently annotated with three types of ground-truth data:
\begin{itemize}
\item A binary risk (risk present or absent), determined via majority vote,
\item A risk type label: one of (cut, fire, fall, or none),
\item A heatmap constructed by averaging participant-marked regions to indicate perceived risk areas.
\end{itemize}
These annotations served as a human-grounded reference for evaluation the output of our algorithm in two dimensions:
\begin{itemize}
    \item \textbf{Semantic accuracy}, i.e., how well the algorithm identifies the correct risk type,
    \item \textbf{Semantic alignment}, i.e., how closely the generated risk heatmaps corresponds to the areas human marked as hazardous. 
\end{itemize}
For both the algorithm and the human annotations, the risk maps were normalized to the range $[0, 1]$ for consistency.

Although this evaluation was conducted using a benchmark dataset, it represents a critical first step in validating our system’s real-world reasoning capabilities. The use of a well-curated dataset enabled controlled comparisons and reproducibility while removing confounding factors such as sensor noise or robot navigation limitations.

A follow-up study is planned to deploy the system on a real robot equipped with an RGB-D camera to assess its effectiveness in dynamic, cluttered, and potentially unstructured household environments. In such future settings, the proposed risk propagation mechanism can serve not only to detect risky configurations in real time but also to proactively guide the robot in mitigating those risks—laying the groundwork for safe and intelligent robotic assistance in domestic contexts.

Such proactive safety behaviors could involve a robot issuing real-time alerts about hazardous object placements, suggesting alternative arrangements, or autonomously intervening to secure unstable items before an accident occurs.


\subsection{Evaluation of Risk Detection Accuracy}
\begin{figure}[tbh]
\centering
\includegraphics[width=0.45\textwidth]{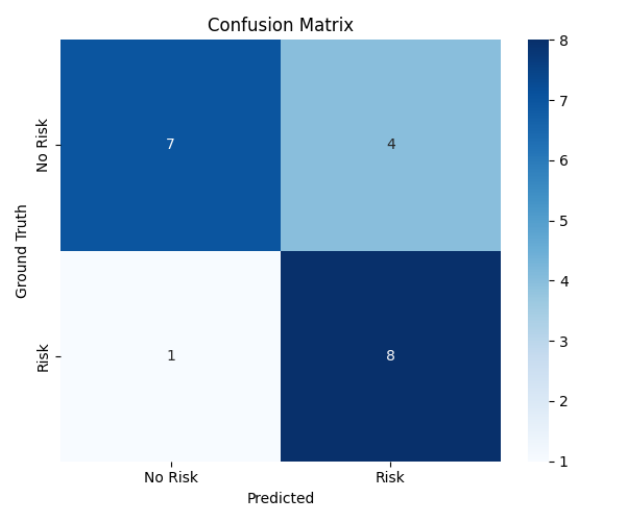} 
\caption{Risk Detection Confusion Matrix}
\label{fig:result01}
\end{figure}
To assess the system’s ability to identify whether an image contains any form of accident risk, we performed a binary classification evaluation using the annotations collected from human participants. Each image in the test set was labeled as either Risk or No Risk, based on participant consensus. These ground-truth labels served as a reference for evaluating the system’s predictions.

An image was predicted as risky if any of the algorithm-generated heatmaps—corresponding to cut, fire, or trip/fall categories—exhibited activation above a predefined threshold. If no such activations were present, the image was labeled as non-risky. This thresholding step ensures that only strongly activated risk regions influence the binary decision.

Figure~\ref{fig:result01} presents the confusion matrix for this binary classification task. The proposed system achieved an overall accuracy of 75\%, demonstrating its effectiveness in correctly identifying the presence or absence of risk in domestic scenes. Notably, the system maintained a relatively low false negative rate, which is critical for safety applications where missed risks can lead to severe consequences.

It is important to note that for this evaluation, predictions were made using only the heatmap corresponding to the risk type annotated for each image. This risk-type-specific evaluation avoids artificial performance inflation that could arise from combining unrelated heatmaps and allows a more consistent assessment of the system’s alignment with human risk perception.

\subsection{Evaluation of Risk Type Classification Accuracy}
\begin{figure}[tbh]
\centering
\includegraphics[width=0.45\textwidth]{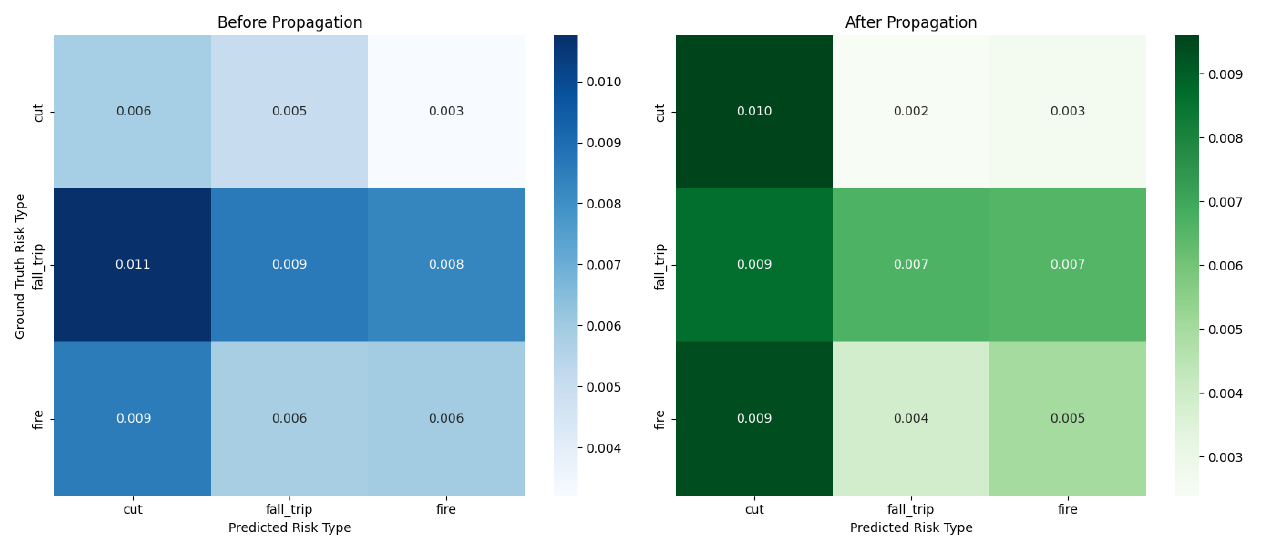}  
\caption{Confusion matrices showing risk type prediction results before (left) and after (right) propagation. The improvement is especially evident in cut-related scenarios.}
\label{fig:result02}
\end{figure}
In addition to detecting whether an image contains risk, we evaluated the system’s ability to correctly classify the type of perceived risk—specifically, whether it corresponds to a cut, fire, or trip/fall hazard.

To assess the spatial alignment between predicted and user-annotated risk regions, we adopted a centroid-based evaluation metric. Unlike traditional Intersection over Union (IoU), which can be sensitive to shape mismatches or boundary noise, this method compares the center of mass of the predicted and ground truth heatmaps. The Euclidean distance d between these centroids is computed as:

\begin{equation}
    d = \|\mathbf{c}_{\text{pred}} - \mathbf{c}_{\text{gt}}\|^2,
\end{equation}

where $\mathbf{c}_{\text{pred}}$ and $\mathbf{c}_{\text{gt}}$ denote the centroid coordinates of the predicted and ground truth heatmaps, respectively.

To interpret this distance as an accuracy-like measure, we use the inverse distance $1/d$, which increases as the predicted centroid approaches the user annotated center. This inverse-distance metric is particularly useful in our context, where heatmaps may differ in shape but still highlight similar risk zones.  A higher value indicates better alignment between predicted and annotated risk regions. It allows a robust comparison even when bounding areas are irregular or non-overlapping.

We performed this evaluation both \textbf{before} and \textbf{after} applying the risk propagation algorithm. The confusion matrices shown in Fig.~\ref{fig:result02} summarize the risk classification results for each scenario. A clear improvement was observed after propagation, especially in cut-related scenes, where localized hazards (e.g., exposed knives or broken glass) benefited from context-aware reasoning.

In parallel, we also evaluated semantic classification accuracy using Intersection over Union (IoU) between the predicted and user-annotated heatmaps. While IoU remains a useful metric for overlap-based validation, the centroid-based method offers a complementary perspective that captures intuitive spatial reasoning.

Together, these evaluations provide a holistic view of the algorithm’s ability to not only detect risk, but also infer its type and spatial extent in alignment with human judgment.

\section{Discussions}
This study aimed to evaluate the feasibility of risk reasoning in domestic environments based on statistical data and object relationships. While the proposed approach demonstrated some capability in identifying risky situations, there remain several limitations that impact overall performance and robustness.

One of the primary causes of false positives in the current system stems from the over-detection of objects in the environment. For example, in a bedroom scene, not only the bed but also the blanket is detected as an independent object. Although the bed alone may suffice to represent the scene, assigning risk scores to every detected item—including relatively benign ones—leads to an overestimation of risk. This phenomenon highlights a limitation of current object detectors when used for safety-critical applications.

Another major challenge lies in the quality and interpretability of the accident data sourced from the governmental databank system. While statistically correct, some entries are counterintuitive from a human perspective. For instance, the system records seven cases of cuts involving "blankets," but these are often indirect or ambiguous in causality (e.g., needles left inside the fabric). Such inconsistencies affect the reliability of the initial risk scores derived for each object.

To mitigate this, incorporating a filtering mechanism based on accident rate thresholds may be beneficial. For example, before calculating the risk score, we can analyze the overall accident share of each object type. If an object accounts for less than a specified percentage (e.g., 0.5\%) of all reported accidents, it may be excluded from propagation altogether. This type of object-level weighting can reduce noise and improve both accuracy and interpretability.

While the proposed risk propagation algorithm effectively identifies cut-related risks, its performance on \textit{fall-trip} and \textit{fire} scenarios was less consistent.

We hypothesize that this is due to the nature of these risk types: 
fall-trip accidents often depend on object states such as pose, orientation, or occlusion, which are not explicitly represented in the current feature set. 
Likewise, fire-related risks require recognition of physical states (e.g., visible flames, heat sources), which are challenging to infer from static object categories alone.
We assigned risk values statistically based on co-occurrence frequencies, but incorporating object-level state recognition (e.g., "lying on the floor", "tipped over", "burning") could significantly improve detection accuracy for these categories.

As discussed above, the current framework, in its nascent stage, lacks critical parameters like comprehensive situational context and human activity. To address this, integrating detailed contextual information via VLM and Semantic Maps. These advancements will enable situational understanding and personalization of perceived risk situations. Ultimately, we plan to implement these frameworks in actual robotic systems and quantitatively evaluate risk reduction in real-world experiments, a crucial step for enhancing safety and reliability in human-robot coexisting environments.

Future work will prioritize improving initial object risk estimation by integrating higher-quality accident data and contextual and developing adaptive thresholding methods to reduce false alarms effectively.

\section{Conclusions}
In this paper, we introduced a novel approach to risk reasoning by modeling accident-prone scenes as relational graphs and propagating risk asymmetrically based on contextual cues. Our method restricts propagation to flow from higher-risk objects to lower-risk ones, guided by accident relevance and distance, ensuring stable convergence and interpretable risk dynamics.

Through comparisons with user-annotated risk maps, we demonstrated that our model achieves 75\% binary classification accuracy, correctly distinguishing risk-present from non-risk scenes. Furthermore, we evaluated the alignment between predicted and ground-truth risk locations, finding that risk propagation improved spatial accuracy in cut-related scenes, though limited gains were observed for fire and fall/trip categories.

These findings suggest that context-aware propagation can meaningfully enhance risk estimation, particularly in scenarios where individual object risk is insufficient. Future work includes refining propagation logic for complex hazard types and expanding the dataset for broader validation.

Importantly, this work lays the groundwork for real-world deployment. While the current evaluation was conducted on static benchmark images, the proposed framework is explicitly designed for integration with robotic systems equipped with RGB-D sensors. A follow-up study will focus on real-time deployment, enabling robots to detect, reason about, and proactively respond to risks in dynamic home environments.

Future extensions will aim to improve robustness by integrating temporal reasoning, multimodal perception (e.g., thermal, motion), and richer object state representations. In addition to these technical directions, a key area of future development is the integration of human presence and behavior into the risk reasoning process. Many potentially hazardous configurations only become truly risky when people are nearby—for instance, a knife left on a table is less concerning when no one is home. Distinguishing between static and person-dependent risks will be essential for reducing false positives and increasing real-world usability. Also while this study focuses on three common accident types, the underlying framework could generalize to broader household risks.

Looking ahead, we envision extending this system into a personalized, household-specific risk mapping framework. Since situations labeled as “risky” may not always align with users' intentions or preferences, it is also important to consider mechanisms for minimizing unnecessary or intrusive interactions, especially in shared living spaces\cite{tafrishi2022psm}. 
For example, a robot might learn that certain areas are intentionally cluttered during specific activities (e.g., crafting) but should be cleared at other times, thus avoiding unnecessary alerts. 

Thus by observing users’ lifestyles and environments over time, a service robot could build and update dynamic risk maps tailored to each home. These maps would capture both static hazards and person-dependent risks, adapting as household conditions evolve.

Such maps could guide the robot’s behavior—prioritizing high-risk areas, planning autonomous patrols, and responding based on situational urgency. For example, it might monitor cluttered walkways in homes with elderly residents or focus on heat sources when children are present\cite{ravankar2020safe}. This would enable robots to proactively address risks, offering timely, personalized support in everyday life.

\section*{Acknowledgment}
This work was partially supported by JST Moonshot R\&D [Grant Number JPMJMS2034] and JSPS Kakenhi [Grant Number JP24K07399].

\bibliographystyle{ieeetr}
\bibliography{main-sena-edits-for-finalsubmission.bib}

\end{document}